\pdfoutput=1
\documentclass[10pt,letterpaper]{article}

\usepackage{cogsci}
\usepackage{amsmath,amssymb}
\cogscifinalcopy % Uncomment this line for the final submission 
\usepackage{graphicx}
\usepackage{booktabs}
\usepackage{siunitx}
\usepackage{tabularx}
\usepackage{multirow}
\usepackage{makecell}
\usepackage[T1]{fontenc}
\usepackage[hidelinks]{hyperref}

\usepackage{tabularray}
\usepackage[table]{xcolor}

\usepackage{pslatex}
\usepackage{apacite}
\usepackage{float} % Roger Levy added this and changed figure/table
\newcommand{\round}[1]{\num[round-mode=places,round-precision=3,]{#1}}

\newcommand{\noserif}[1]{{\scriptsize \fontfamily{phv}\selectfont #1}}
\newcommand{\nosbold}[1]{\textbf{ \fontfamily{phv}\selectfont #1}}
\newcommand{\acc}[2]{\tiny \round{#1} \tiny ${\scriptstyle \pm}$\round{#2}}

\newcommand{\gacc}[2]{\graycellcolor\round{#1}&\graycellcolor\ifthenelse{\equal{#2}{}}{}{\tiny ${\scriptstyle \pm}$\round{#2}}}

\newcommand{\ACC}[2]{\textbf{\round{#1}}&\ifthenelse{\equal{#2}{}}{}{\textbf{\tiny ${\scriptstyle \pm}$\round{#2}}}}

\usepackage{color}

\title{Representations as Language: An Information-Theoretic Framework for Interpretability}

  \author{
    \large{ \bf Henry Conklin$^{\bullet, \circ}$, Kenny Smith$^{\circ}$}\\
    \emph{\{henry.conklin, kenny.smith\}@ed.ac.uk} \\
    $^{\bullet}$Institute for Language Cognition and Computation, School of Informatics, The University of Edinburgh\\
    $^{\circ}$Centre for Language Evolution, The University of Edinburgh}

\begin{document}

\maketitle

\begin{abstract}

  Large scale neural models show impressive performance across a wide array of linguistic tasks. Despite this they remain, largely, black-boxes - inducing vector-representations of their input that prove difficult to interpret. This limits our ability to understand what they learn, and when the learn it, or describe what kinds of representations generalise well out of distribution. To address this we introduce a novel approach to interpretability that looks at the mapping a model learns from sentences to representations as a kind of language in its own right. In doing so we introduce a set of information-theoretic measures that quantify how structured a model's representations are with respect to its input, and when during training that structure arises. Our measures are fast to compute, grounded in linguistic theory, and can predict which models will generalise best based on their representations. We use these measures to describe two distinct phases of training a transformer: an initial phase of in-distribution learning which reduces task loss, then a second stage where representations becoming robust to noise. Generalisation performance begins to increase during this second phase, drawing a link between generalisation and robustness to noise. Finally we look at how model size affects the structure of the representational space, showing that larger models ultimately compress their representations more than their smaller counterparts\footnote{Code and additional visuals at \href{https://www.github.com/hcoxec/h}{\texttt{github.com/hcoxec/h}}}.

%These measures predict which representations will generalise more robustly.

%we retain limited ability to determine how they do so and how they represent their training data. We introduce an information-theoretic approach to interpreting what large-scale transformer models learn, and when. Characterising the system-level structure that emerges in a model's representations over the course of training. This enables us to 1) describe 3 distinct phases of training, 2) quantify the degree of linguistic structure in a model's mappings from inputs to representations, 3) predict which models will generalise best based solely on their representations for two different datasets designed to evaluate compositional generalisation.

\end{abstract}

\section{Introduction}

Deep-Learning models achieve remarkable performance across a broad range of natural-language tasks \cite{vaswani_attention_2017}, but we still have a limited understanding of the learning process they undertake, and how they come to represent information so effectively. This is in part because these models are black-boxes \cite{tishby_deep_2015, shwartz-ziv_opening_2017}. They learn representations of their training data that are high-dimensional vectors, gigantic lists of numbers that are hard to interpret. While there is a growing body of work on interpretability, offering techniques for predicting what is encoded in a model's representations \cite{voita_information-theoretic_2020, pimentel2020information}, there's still lack of clarity about how representations themselves are structured, how that structure emerges, and what kinds of structures are desirable. 

\begin{figure}
    \centering
    \includegraphics[width=0.5\textwidth]{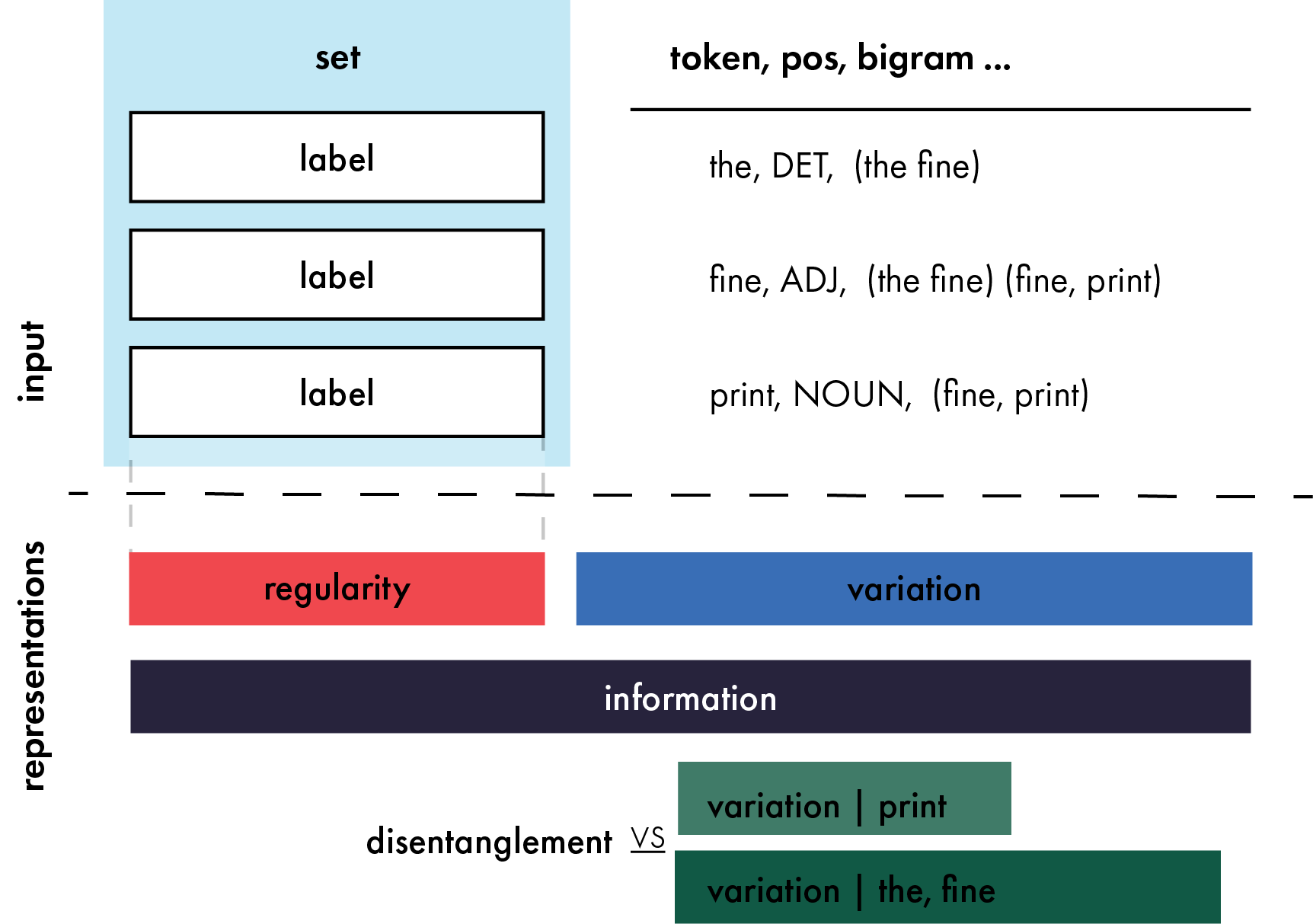}
    
    \label{fig:basic_overview}
    \caption{a depiction of basic quantities we measure and how they relate to each other. We measure structure in the mapping between labels for the dataset, and latent representations inside of a transformer. Here some example labels are given for the sentence "the fine print."}
\end{figure}

Central to language's ability to generalise is its regularity, exemplified by syntactic structure \cite{partee1995lexical}, which allows predictable \& regular encoding of meanings across the entire system. Languages are also rich with variation which can make them more expressive \cite{hurford2003synonymy} and structured ambiguities that can make them more compressible \cite{piantadosi_communicative_2012}.
Do the representations learned by a transformer model \cite{vaswani_attention_2017} exhibit similar \emph{system level-structures}? To answer this we look at the representations that emerge over the course of training as a kind of language in their own right.  At a high-level we can think of language as a mapping between spaces, like between meaning and form \cite{saussure_course_1916}. A multi-layered neural model needs to learn to map a sentence to a vector representation that later layers can successfully map to the output; encoder-decoder models \cite{cho2014learning} even more explicitly use separate parts of a model to map in and out of vector space. We draw an analogy between these two mappings, quantifying different kinds of regularity and variation in a model's mapping between inputs and representations. While there has long been interest in the kinds of representations learned by deep-models \cite{bengio2013representation, locatello2019challenging}, there has been little work quantifying systematic structure in the representations learned by transformers or relating them to the kinds of structures that characterise natural language. It's worth noting our approach is in contrast to some existing work that draws parallels between model weights and formal languages \cite<trying to infer functions or 'source code' from model weights; >{elhage2021mathematical} --- we think an approach grounded in natural language is more scalable and better suited to characterising the kinds of systematic structure \& variation that emerge in deep-learning models, especially those trained on data from natural languages. 

We introduce a novel information-theoretic framework for assessing whether the representations learned by a model are systematic. In order to do this we first discretise vector representations into a sequence of symbols, then quantify 4 properties of the learned mapping from sentences to symbols: the degree of compression, regularity, variation and disentanglement. By doing this at different levels of abstraction we show when lexical and syntactic information are learned. We can identify two clear phases of training, the first characterised by the model rapidly learning to disentangle and align representations with token and part of speech information, the second (far longer) phase of training characterised by representations becoming more robust to noise. During this second phase models compress their representations, with larger models compressing considerably more; at the same time, generalisation performance begins to slowly improve, showing a link between robustness to noise and generalisation. Finally we discuss what kinds of representational structure are desirable, using our measures to predict which models will perform best on a generalisation set.

\section{Methods}

Our experiments use a Transformer \cite{vaswani_attention_2017} encoder-decoder model, with a two layer encoder, and single layer decoder. The model's encoder maps each input sentence to a vector representation, then the decoder uses this representation to generate the corresponding output, in our case a formal semantic representation of the input sentence. We look at the encoder's mapping between sentences and representations, and quantify the degree to which it exhibits systematic structure. We train each model (from scratch) on two different semantic parsing datasets, designed to evaluate a model's ability to systematically generalise: SLOG \cite{li_slog_2023} where the task is to generate lambda expressions for a sentence, and CFQ-MCD \cite{keysers_measuring_2020} where questions about movies need to be mapped to SQL queries that answer them. Both of these datasets come with an out-of-distribution generalisation set containing examples generated by the same grammar as the training data, but purposefully designed to be challenging. We also look at whether the capacity of a model affects the kinds of structure that emerges, training three different model sizes (with either 64, 128, or 256 hidden dimensions).

\subsection{Estimating Entropy in Vector space}

Shannon Entropy describes the amount of information contained in a random variable \cite{Shannon1948AMT}. While methods exist for estimating entropy of continuous variables, these approaches are difficult to compare across representational spaces and often require strong assumptions about the underlying probability distribution \cite{Jaynes1957InformationTA}. Instead we discretise the hidden representations into a sequence of random variables, enabling us to directly estimate the Shannon entropy of our latent space. Our method is analogous to converting each vector into a sequence of discrete symbols, with a symbol for each dimension of vector space. Previous information theoretic analyses of deep learning have performed a similar estimation (e.g. \citeNP{saxe2019information}), although it's been noted this approach is more reflective of non-uniformity, like clustering behaviour, than it is of the true entropy of the space \cite{goldfeld2018estimating}. For our purposes identifying the degree to which a variable is uniformly distributed, or tightly clustered is sufficient to draw substantive conclusions.

For a given vector in a set of vectors $v_i \in V$ with dimensions $d \in D$ we cut each dimension $V_d$, into $N$ equal-width bins between the attested maximum and minimum values of that $V_d$. This enables a straightforward maximum-likelihood estimate of the entropy of $V$ by counting the frequency of each bin and normalising by number of representations in $V$. Resulting bin probabilities $p(V_{dn})$ are used to estimate the entropy of each dimension, then averaged across dimensions to give us an overall estimate of the dimension-wise entropy of $V$. 

\begin{equation}
    \label{eq:H}
    H_{dw}(V) = \frac{1}{|D|}\sum_{d}^{D}\sum_{n}^{N} -p(V_{dn})\log(p(V_{dn}))
\end{equation}

\noindent On the right in \ref{eq:H} is the equation for shannon entropy \cite{Shannon1948AMT}. As this estimate is an approximation we also use the Miller-Meadow correction in order to smooth the estimate based on sample size and improve its accuracy \cite{Miller1955NoteOT}. No method of estimating discrete entropy in continuous spaces is perfect (see \citeNP{paninski_estimation_2003} for extensive discussion), but our estimator is invariant to linear transformations while making minimal assumptions about the underlying distribution. Note that while in the results presented here we estimate entropy per dimension we can just as easily estimate entropy per pair or set of dimensions (akin to modelling at the unigram vs n-gram level); our use of a dimension-wise estimate simplifies our analysis but limits its ability to track cross-dimensional dependencies. Additionally it allows us some insight into the role of different subspaces of the representational space, by letting us break the estimate down dimension by dimension.

\subsection{Measuring Structure}\label{measures}
We are interested in whether a model's representations become systematically structured during training, reflecting the system-level structure of the data they're trained on. Using our entropy estimator we can assess 4 different quantities at different levels of abstraction, which allow us to describe the degree to which the representations a model learns are structured with respect to structure in the dataset it's trained on. Here we walk through our measures for describing the representational system that emerges over the course of training, quantifying the amount of Information, Variation, Regularity, and Disentanglement. 

\paragraph{Information:} We have a model $f$ that maps a set of sentences $X$ to representational space $Y$. For each sentence $S^k \in X$, the model takes as input a sequence of tokens --- usually words --- $t_a^k, t_b^k, t_c^k ... \in S^k$ and returns a sequence of vectors $v_a^k, v_b^k, v_c^k ... \in V^k$ where $v_a^k$ is the vector corresponding to token $a$ when it occurs in sentence $k$. While each sequence $V^k$ is of variable length, the individual vectors are the same size. We can therefore create a list $Y$ of all token representations from all sentences in the dataset

\begin{equation}
    Y = [v_a^k : \forall v_a^k \in f(S^k) : \forall S^k \in X]
\end{equation}

\noindent and calculate its dimension-wise entropy. The result gives us a measure of the average amount of information encoded in each dimension of the representation, $H_{dw}(Y)$. Given that the amount of information the model needs to encode is constant (the dataset doesn't change during training) this also tells us how compressed the model's representations are. As the dimension-wise entropy goes down, the model uses less of its available representational space. Information is minimised (i.e. compression is maximised) as all tokens are mapped to the same vector regardless of the token and sentence they correspond to, and information is maximised when token representations are spread out uniformly across representational space. To aid interpretation we normalise this measure, as well as Variation and Regularity, so that 1.0 indicates a uniform distribution and 0.0 is one-hot.
Our estimator is \emph{invariant to linear transformations,} which means it ignores how numerically large a representational space is used. That is, this score is maximised if representations are spread out uniformly between the interval -2, 2 or -10,10 --- what matters is representations' dispersion, not their magnitude.

\paragraph{Variation} captures how much a property varies in representation space. Given a class of labels, like tokens, or parts of speech, it reflects whether the model learns a single global representation of each label invariant to context, or if each representation is completely unique to the sentence it occurs in. We quantify this in terms of the conditional entropy of representations, given a label, creating a list of all instances of that label $Y|label$, across all contexts where it occurs

\begin{subequations}
\begin{equation}
    Y|label = [v_a^k \text{ \emph{if} } a=label : \forall v_a^k \in Y]
\end{equation}

\noindent Labels for the tokens fed into a model are virtually always known, so we can easily estimate the conditional dimension-wise entropy of $Y$ given a specific token $H_{dw}(Y|token)$. This is minimised when all instances of a token map to the same vector regardless of the sentence they occur in, and maximised when $H_{dw}(Y|token) = H_{dw}(Y)$ indicating instances of the same token are no more likely to be similar than two tokens chosen at random. The mean variation across the set $S$ of all tokens gives us a general sense of how much the model encodes context in its internal representations.

\begin{equation}
    variation(Y|Set) = \frac{1}{|S|}\sum_{label}^{S}H_{dw}(Y|label)
\end{equation}

\noindent We can also calculate variation with respect to any features we have a set of labels for. For example, if we know the part of speech for each of the input tokens $variation(Y|POS)$ could tell us if members of the same syntactic class share more information with each other than expected by chance. In the general case we just need a set of labels to condition on (e.g. part of speech, morphological case, tense etc.) when estimating $H_{dw}(Y|Set)$.

\end{subequations}

\paragraph{Regularity} measures how structured a model's representations are with respect to a feature in the input --- in particular, whether the mapping between a label and its representation is monotonic (one-to-one).  The inverse of variation, Regularity quantifies how much knowing something about a token is going to tell us about its representation; quantifiable as the dimension-wise mutual information between a label and its representations.

\begin{equation}
    regularity(Y, Set) = \frac{1}{|S|}\sum^{S}_{label} H_{dw}(Y) - H_{dw}(Y|label)
\end{equation}

\noindent This is maximised when a label and its representations are monotonically aligned --- knowing the label tells us everything there is to know about the representation. As with variation we can quantify regularity with respect to individual labels in a set and mean across them to get a general notion of how aligned representations are with e.g. tokens, POS tags, or the bigrams a token is part of.

\begin{figure}

    \includegraphics[width=0.45\textwidth]{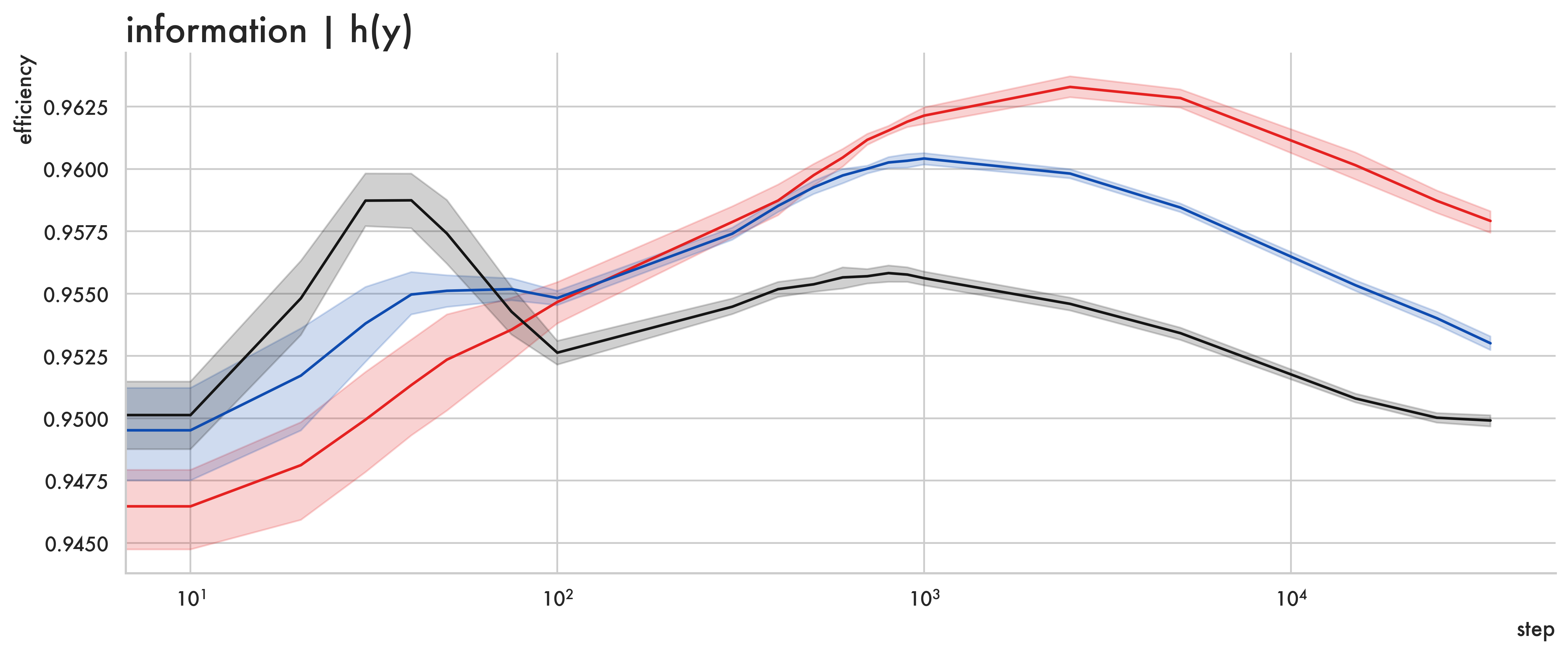}
    \includegraphics[width=0.45\textwidth]{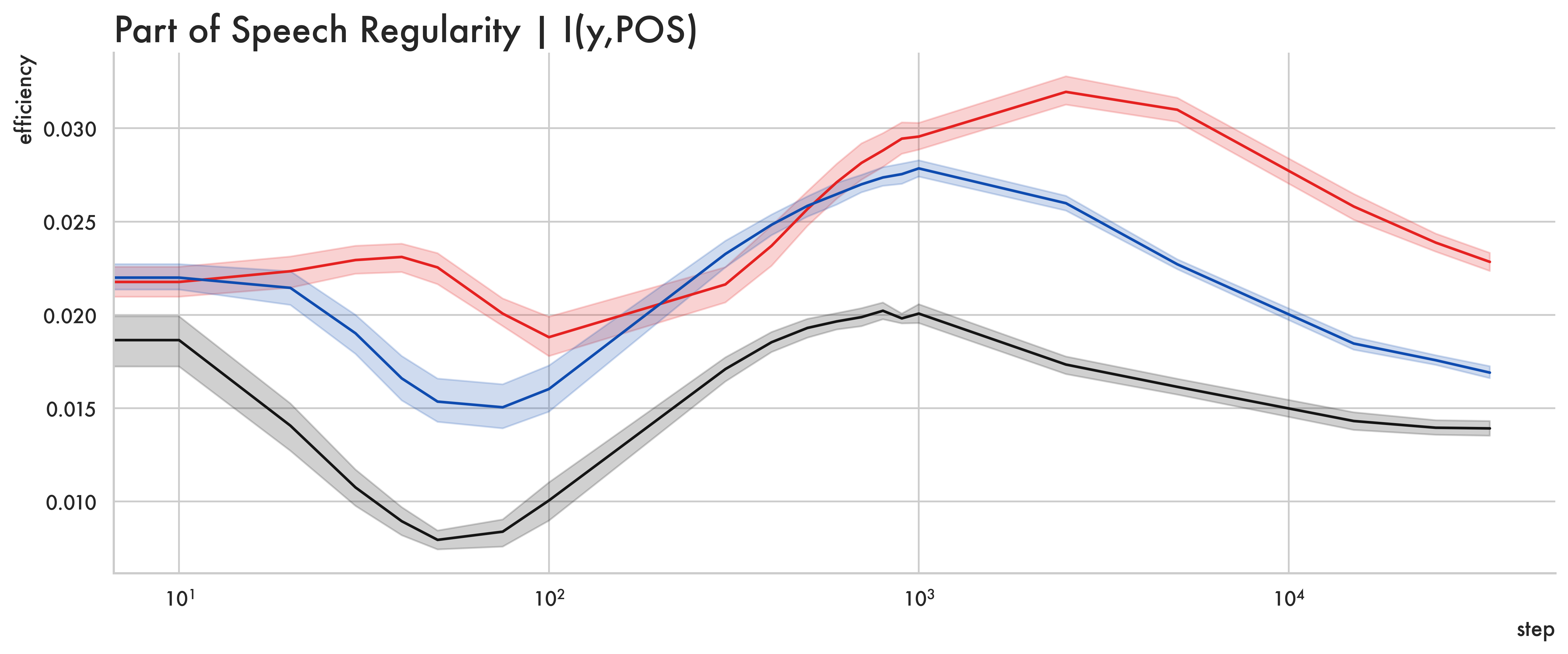}
    \includegraphics[width=0.45\textwidth]{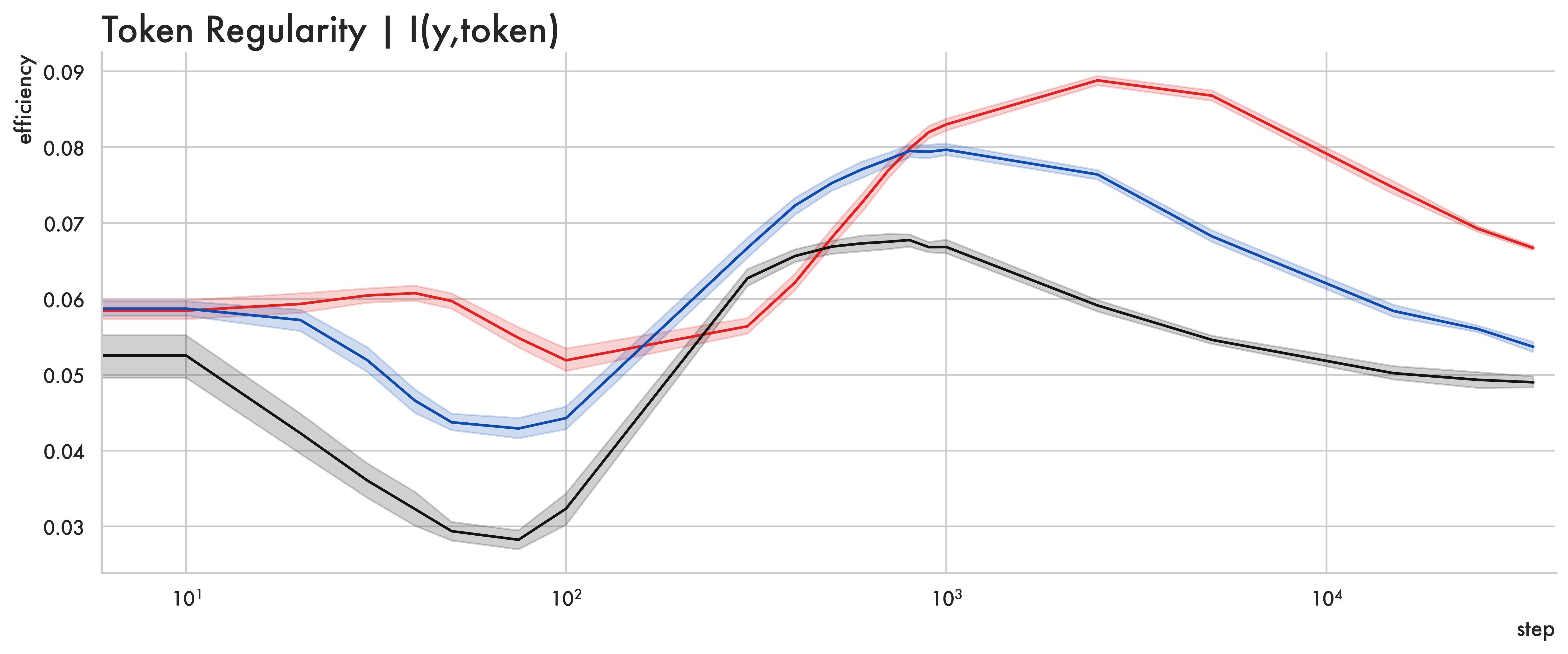}
    \includegraphics[width=0.45\textwidth]{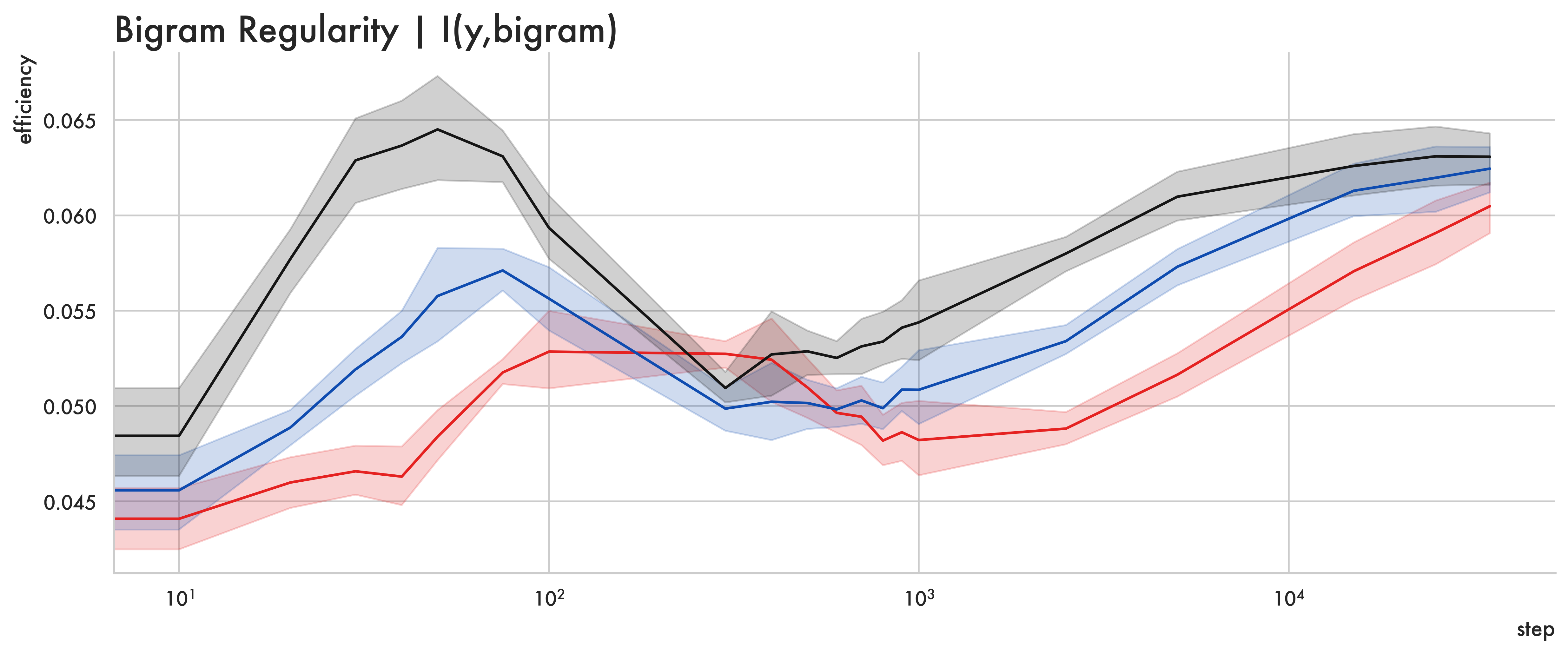}

    \includegraphics[width=0.45\textwidth]{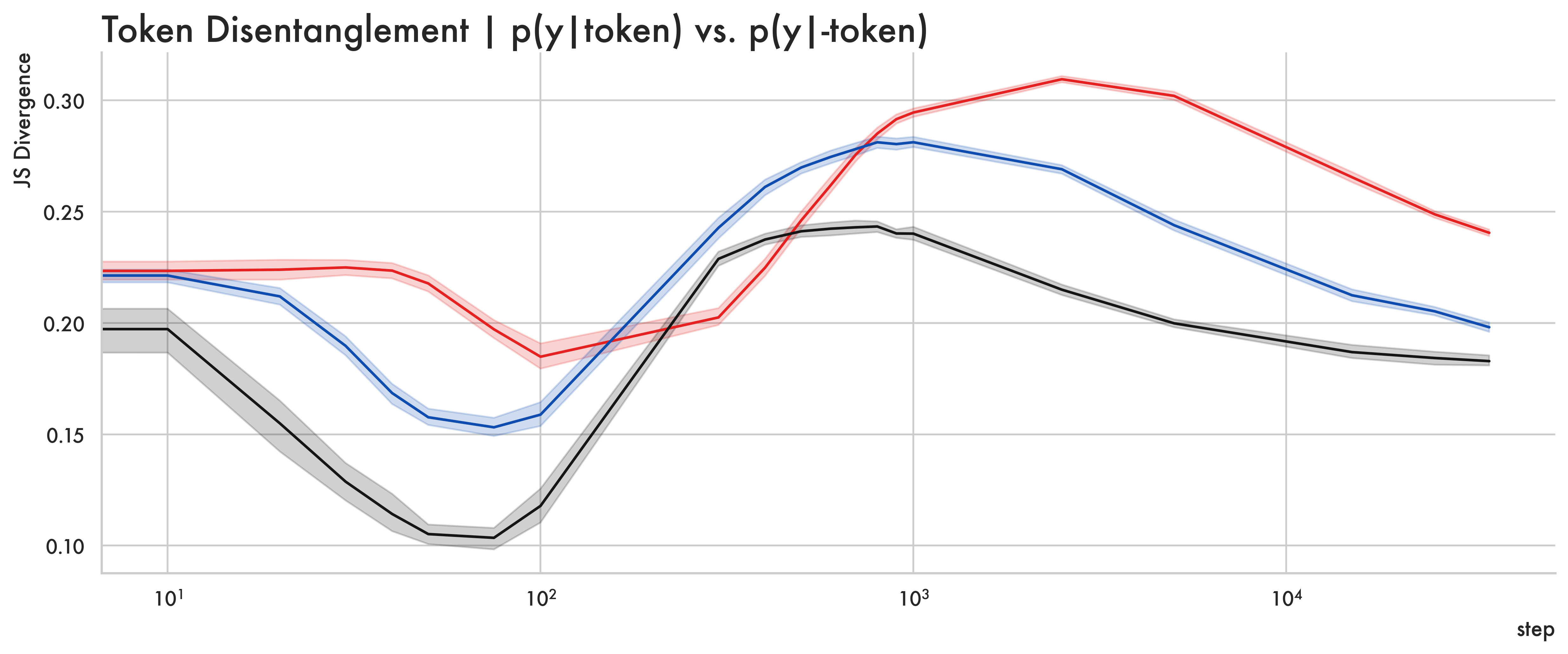}
    \includegraphics[width=0.45\textwidth]{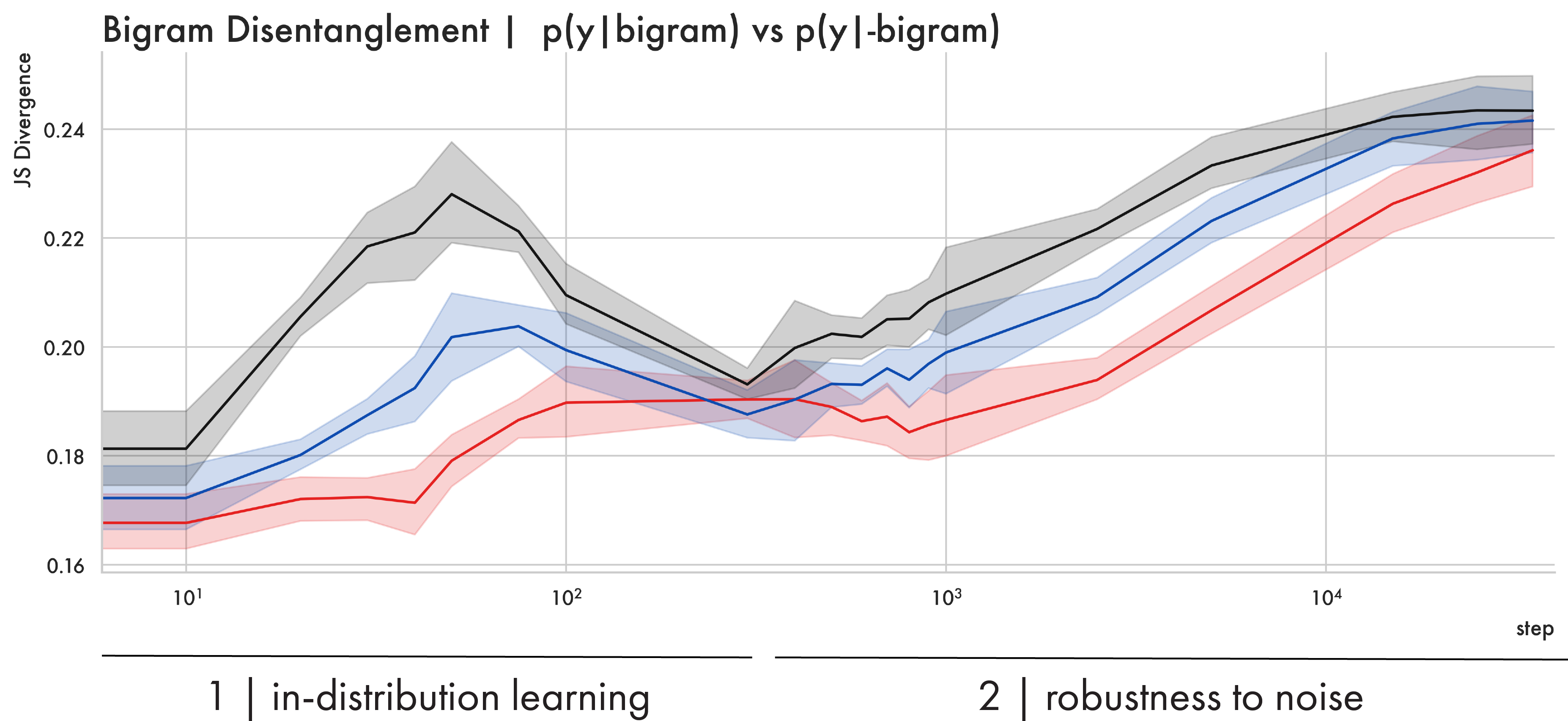}

    \caption[short]{Each facet shows a different measure (along the y axis) against training steps (log scaled). Lines and shading give means and 95\% CIs; line colours give results for 3 different model sizes. Values are calculated across the entire training set for 10 different random seeds. Efficiency (normalised entropy) is bounded  such that 1.0 indicates a uniform distribution and 0.0 one-hot.}

    \label{fig:slog_sizes}
\end{figure}

\begin{figure}
    \centering
    
    \includegraphics[width=0.45\textwidth]{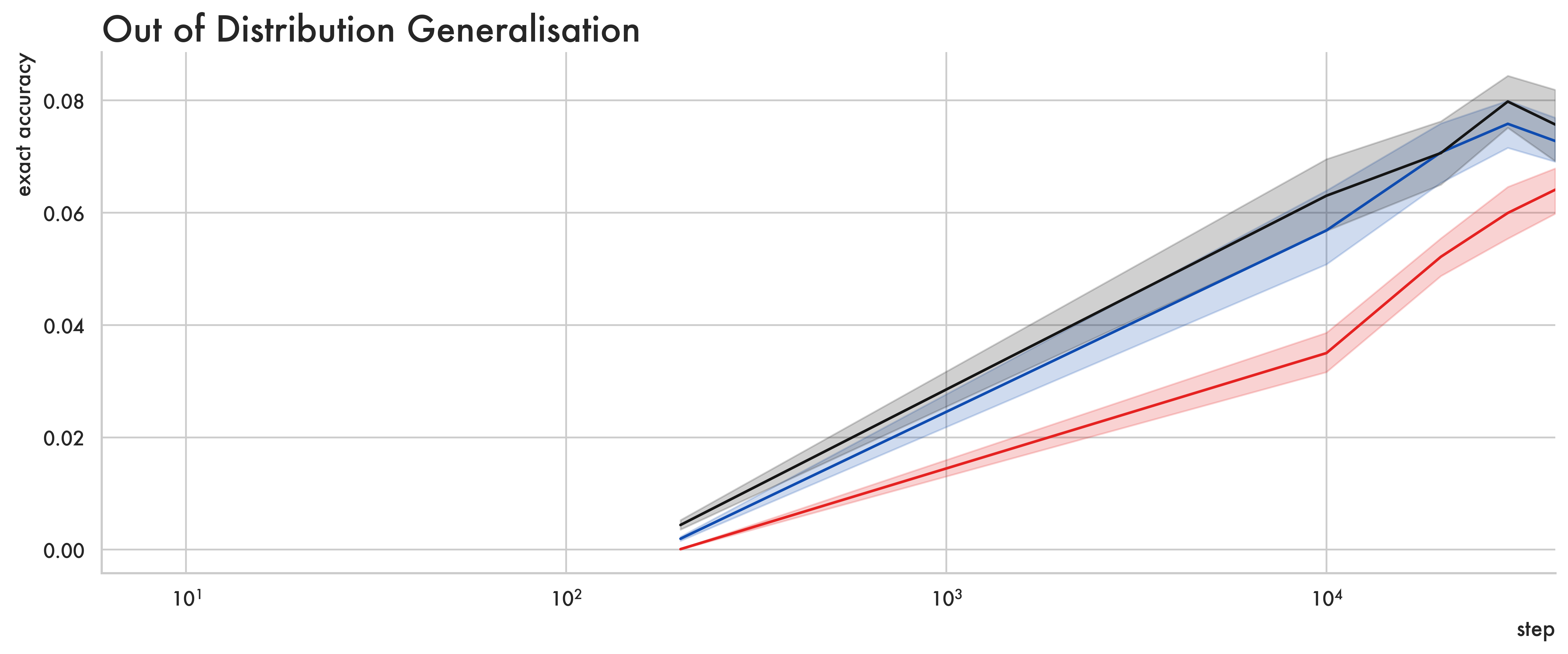}
    \includegraphics[width=0.4\textwidth]{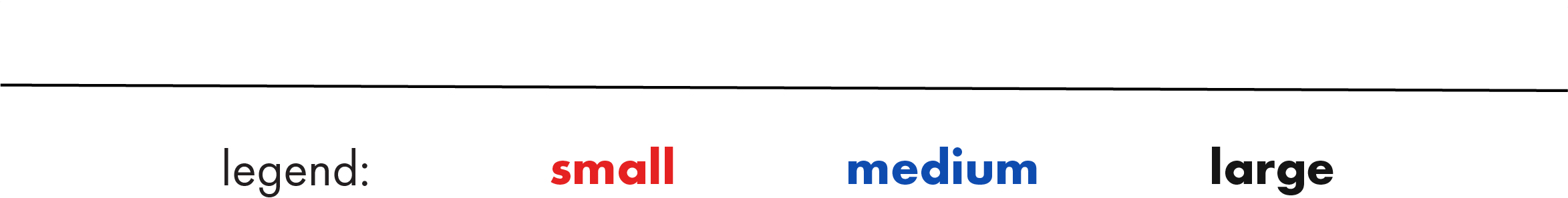}

    \label{fig:acc}

\end{figure}

\paragraph{Disentanglement} measures how separable different labels within a set are from one another, e.g. whether separate tokens are represented in distinct regions of representational space, rather than overlapping. We measure this by assessing the Jensen-Shannon divergence between $P(Y|label)$ and all other labels in the set $P(Y|Set \neg label)$; if tokens are distributed uniformly across a space their disentanglement will be 0, while if they are entirely separable it approaches 1.

\begin{equation}
    dis(Y, Set) = JSD(P(Y|label) ; P(Y|Set \neg label))
\end{equation}

\noindent As with previous measures we aggregate this to get an assessment of how disentangled the class of labels is. This measure is related to previous assessments of entanglement \cite{chen2018isolating, conklin2022compositionality} but is implemented quite differently, and requires no pair-wise comparison of different labels.

\section{Results}

We report results on two different datasets designed to assess compositional generalisation. Our measures allow us to characterise the trajectory of training, which we identify as having two distinct phases. We also compare models of different sizes to see how capacity changes representational space. Summary results are found in table \ref{tab:main_results}. 
It is worth noting that some of our results may be particular to the hyperparameters used for training. We use hyperparameters recommended by the authors of the datasets we use, or that the transformer was introduced with \citeA{vaswani_attention_2017}. We believe this means that our design choices are representative of common ones for training sequence-to-sequence transformers, our code itemises all parameters used. We compute measures with respect to labels for Tokens, Parts of Speech, and Bigrams in the input and for brevity report the values with the clearest effect on model performance. We also focus discussion on results from the MCD CFQ dataset, as it's the larger of the two (100,000 training examples) and is a more realistic task --- mapping questions to SQL queries. We report results for the most challenging split of this dataset, known as MCD2. We include some discussion of SLOG, but an exhaustive listing of all results, across all datasets levels of analysis and model sizes can be found with the released code. 

\subsection{Two Distinct Phases of Training}
We see 2 distinct phases of training, similarly to \citeA{shwartz-ziv_opening_2017} who also describes two phases of training, despite using rather different methods (studying classification with a feed-forward network rather than a linguistic task with a transformer). This suggests some generality to this characterisation of deep-learning, though our results point to different analyses of each phase (particularly the second, much longer one), likely due in large part to the difference in model and domain. While overall trajectories are consistent across conditions when different model sizes move between phases differs, for clarity here we refer to specific steps in the training timeline for the mid-size model on CFQ.

\subsubsection{Phase 1 | In-Distribution Learning}
:\newline
\emph{Alignment \& Disentanglement.}
In Phase 1 the model achieves high in-distribution accuracy, climbing to ceiling performance on the training data by step 1,000. This increase in accuracy is driven by an increase in token and POS regularity between steps 100 and 1000 as representations become more monotonically aligned with the corresponding input token and its part of speech (fig ~\ref{fig:slog_sizes}, top left). This period also reflects an increase in POS and Token disentanglement indicating different tokens are represented in increasingly distinct regions of representational space. Conversely bigram regularity and disentanglement are reduced over the same interval as different token representations in the same bigram become more uniformly distributed over the support of $Y|token$.

\begin{figure}[t]
    
    \includegraphics[width=0.5\textwidth]{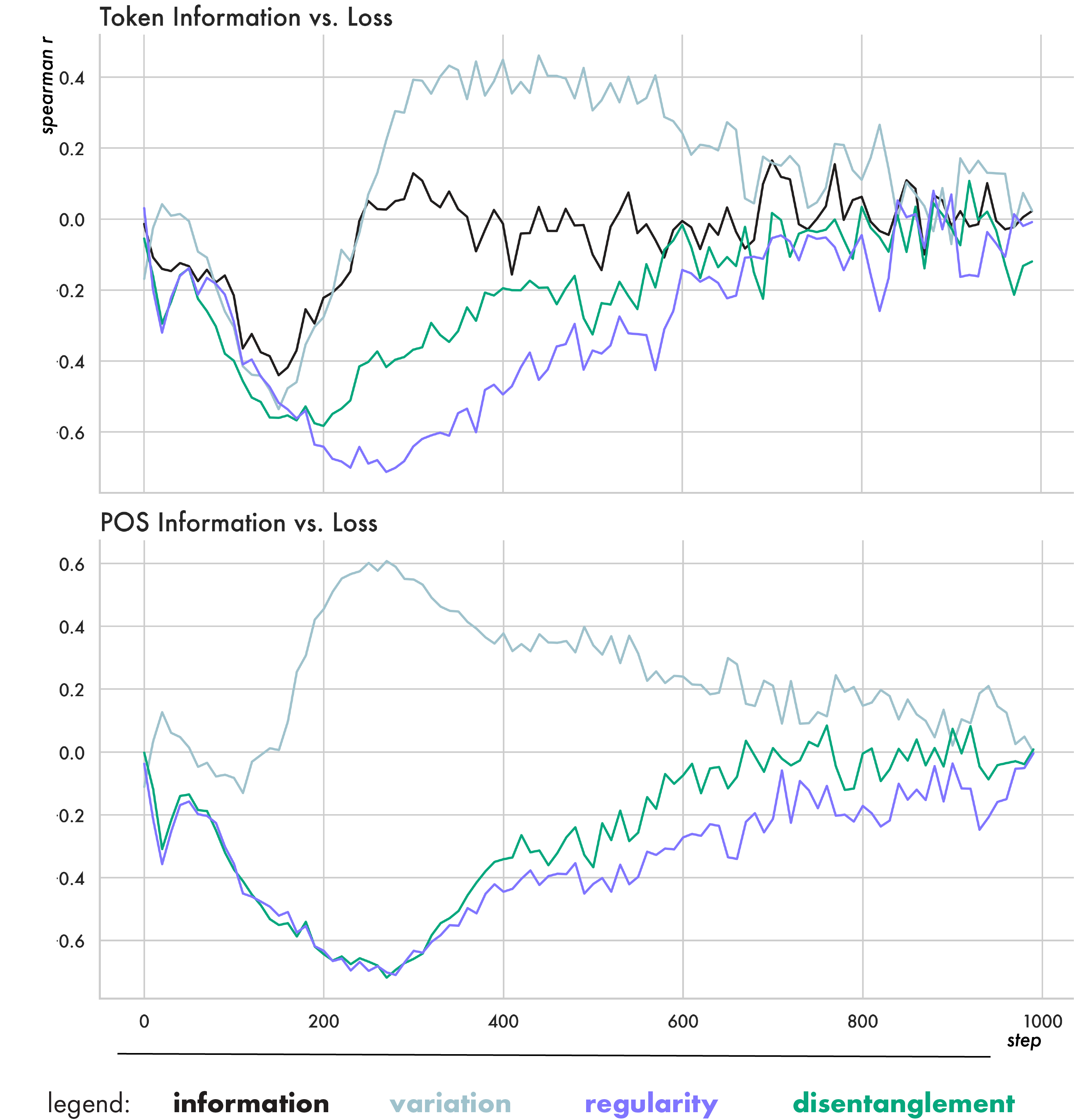}

    \caption{Spearman correlation coefficients between the loss minimised during training and our measures. Negative coefficients suggest the objective increases a quantity. Results for 100 runs of the medium model on SLOG. When exactly significance fades is noted in body text.  
    \textbf{Top:} measures conditioned on token labels. \textbf{Bottom:} measures conditioned on part of speech tags for the tokens.
    }
    \label{fig:loss_corr}
\end{figure}

\paragraph{Does training select for structure?} During training the model tries to minimise a loss function, here the cross entropy between its predicted semantic representation for the input sentence and the correct one. During phase 1 we find a lower loss on the task (indicating better performance) correlates with our measures, suggesting the objective selects for certain structural properties in representation space. The timecourse of this is shown in figure ~\ref{fig:loss_corr}, with correlations between structure measures and the loss for 100 different runs of the medium model on SLOG. From steps 100 to 200 all four of our token-level measures correlate negatively with task loss ($p<0.001$), 
This dynamic shifts slightly from steps 200-600, where higher token disentanglement ($p<0.001$) and regularity (indicative of a more monotonic alignment) ($p<0.001$) continue to correlate with lower task loss but now with less variation ($p<0.001$, steps 280-600). 

%\definecolor{tableShade}{blue}{0.97}

\begin{table}[t]
    \centering
    \rowcolors{3}{white}{blue!3}
    \begin{tabularx}{\linewidth}{l|XXX}                        
    
    \nosbold{SLOG} & \noserif{POS} & \noserif{Token} & \noserif{Bigram} \\
    \midrule
    
    \scriptsize{Information} & \acc{0.956647}{0.000303}& \acc{0.956647}{0.000303} & \acc{0.956647}{0.000303} \\
    \scriptsize{Variation} & \acc{0.920196}{0.000784} & \acc{0.849814}{0.000624} & \acc{0.754133}{0.000448} \\
    \scriptsize{Regularity} & \acc{0.036451}{0.000624} & \acc{0.106833}{0.000762} & \acc{0.096951}{0.000757}\\
    \scriptsize{Disentanglement} & \acc{0.147554}{0.002214} & \acc{0.353022}{0.001694} & \acc{0.373025}{0.001775} \\

    \multicolumn{4}{r}{\scriptsize{Accuracy:} \acc{27.8432}{0.7613}} \\

    \midrule
    \multicolumn{4}{l}{\nosbold{CFQ}} \\
    \midrule
    \scriptsize{Information} & \acc{0.952721}{0.000240} & \acc{0.952721}{0.000240} & \acc{0.952721}{0.000240} \\
    \scriptsize{Variation} & \acc{0.936019}{0.000327} & \acc{0.899654}{0.000468} & \acc{0.836738}{0.001703} \\
    \scriptsize{Regularity} & \acc{0.016702}{0.000292} & \acc{0.053066}{0.000540} & \acc{0.063160}{0.001236} \\
    
    \scriptsize{Disentanglement} & \acc{0.070673}{0.001215} & \acc{0.196321}{0.001453} & \acc{0.241247}{0.004246} \\

    \multicolumn{4}{r}{\scriptsize{Accuracy:} \acc{7.5936}{0.4130}} \\ 
    
    \bottomrule
  \end{tabularx}
  \caption{Summary results for measures at the POS, Token, and Bigram level, across 10 runs of the medium model on both datasets, with 95\%CIs. Measures are computed at the last step of training across the entire training set. Models' accuracy \% reported on the held out generalisation set. 
  %For CFQ we only report results for the most challenging split (MCD2)
  }
  \label{tab:main_results}
\end{table}

Past this point all measures cease correlating with the task loss, which is also the point where empirical error begins to saturate --- as the model approaches ceiling performance on the training set, the loss asymptotically approaches its floor. Figure ~\ref{fig:loss_corr} also shows the correlation between loss and our measures conditioned on part of speech tags. Similarly, greater regularity and disentanglement with respect to part of speech labels and less variation correlate strongly with a better task loss from step 100 until 600 ($p<0.001$). The peak spearman coefficient for disentanglement reaches -0.71 indicating the objective optimizes more strongly for disentanglement of parts of speech than tokens (which peaks at -0.58 and fades from significance faster).

\subsubsection{Phase 2 | Robustness to Noise}:\newline
\emph{Contextualisation \& Compression}
 This is the dominant dynamic of training, taking place from step 1000 onwards. During this period the representational space slowly compresses, with dimension-wise entropy decreasing. This is coupled with an increase in bigram regularity as clusters for different contextualizations become more distinct in representational space, forcing the overall token regularity down. These shifts happen slowly taking thousands of training steps.

\citeA{shwartz-ziv_opening_2017} note that later in training, after the loss has reached its floor, the update steps the model takes begin to behave like `gaussian noise with very small means.' This aligns with what we see here, as measures of structure cease to consistently correlate with the task objective by phase 2. This suggests that a major dynamic of the latter period of training is representations becoming increasingly robust to noise. The model's mapping from sentences to representations needs to continue to encode the input, but do so robustly enough that the mapping won't be undermined by constant noisy updates, otherwise the task loss will begin to increase. Unlike previous work we note mutual information increases between inputs and representations later in training, just at higher level of granularity --- here, bigrams. 

 It's also worth noting that while the model achieves ceiling performance on the training and validation data during phase 1, it only begins to succeed on the more challenging out-of-distribution generalisation task 10,000 steps later (see figure \ref{fig:slog_sizes} top right). This means robust generalisation ability begins to appear only after a sustained period of representations becoming more robust to noise. This is related to the double descent phenomenon \cite{nakkiran2021deep}, where models begin to exhibit strong generalisation performance long after the initial learning of in-distribution data. \citeA{voita_language_2021} also note that in machine translation a transformer starts by learning individual token probabilities before acquiring more complex sentential structure. Our results give a mechanistic account of how this may happen, with token alignment increasing first, then a much longer phase where representations become more contextualised. Though our task is simpler than large-scale translation, in future we aim to apply this analysis to that context.

\paragraph{What kinds of representations generalise best?}
We also look within conditions to see if representational structure correlates with generalisation across different runs of the same model. We take the middle-sized model on CFQ and correlate across 10 runs at the final step of training. This analysis shows that runs with higher bigram disentanglement ($r=0.65$, $p=0.04$), and higher bigram regularity generalise better ($r=0.61$, $p=0.06$). The generalisation set of CFQ contains tokens seen during training as part of novel contexts. In order to do well our model needs to correctly encode tokens it has seen before, in contexts it hasn't. Higher bigram regularity and disentanglement indicates different contextualisations for the same token are more tightly clustered in space and that those clusters are more pure, which may help novel contextualisations of a token to be decoded correctly.
 
\subsection{Model Size Clearly Affects Representational Space}
While the overall phases of training are remarkably consistent across datasets and model sizes, there is a clear influence of model size on representational structure. Figure \ref{fig:slog_sizes} shows trajectories for our three different model sizes over the course of training. Smaller models are less compressed, and have greater regularity and disentanglement with respect to tokens and parts of speech. They also perform worse on both tasks than their larger counterparts. Larger models are more entangled at the POS and token level, but have more disentangled bigrams --- indicating larger models learn more pure clusters for different contextualisations of the same token.

\paragraph{Why Models Compress \& Larger Models Compress More}
It's common to think of connectionist models as cognitive models, and expect them to be governed by similar constraints \cite{futrell2018rnns}. Humans may generalise robustly because constraints on our cognitive capacity force us to learn generalisable solutions rather than memorizing every possible outcome \cite{Griffiths2020UnderstandingHI, hahn_resource-rational_2022}. The fact that larger models (with greater capacity) compress more per-dimension, would seem at odds with this framing. 
While we agree that drawing cognitive parallels can be useful, on a representational level looking at models as a language can help us to reason about the effects of scale and the phases of training.

Specifically, our interpretation is that larger models are able to exploit their higher-dimension internal representations to develop representations more robust to noise. An obvious analogy in communication is mapping an input to a discrete signal, where the signal space is defined by an alphabet of characters and a maximal signal length. If the signal length is low, a larger alphabet is needed to encode the input unambiguously. In contrast, if longer signals are allowed a smaller alphabet is required, the limiting case being a binary alphabet (like morse code) where sentences are encoded in comparatively long signals. Signals composed from a smaller alphabet are more resilient to noise \footnote{This is implicit in Shannon's definition of entropy, as the maximum uncertainty of a binary distribution is lower than one with more outcomes ($log(2)<log(3)$)} for instance, when an operator interprets morse code, at each point in the sequence they only need to differentiate between two possibilities, dot or dash, which is easier than distinguishing between e.g. 26 different outcomes, particularly on signals transmitted over copper wire. We have shown how, during the second phase of training, transformers compress their representations in response to noisy update steps. This is directly analogous to models using a progressively smaller vocabulary for each dimension of hidden space. Larger models have more dimensions, which in our analysis is akin to having a longer maximum signal length, enabling them to learn a mapping more robust to noise, like morse code, converging to a smaller alphabet but longer signal. 

\section{Conclusion}
We have introduced a linguistically-motivated approach to interpreting transformer models. By looking for system-level structure in the model's representations, we characterise two-distinct phases of training, and show how representational structure develops during those phases and how this explains model's ability to generalise. This is enabled by an efficient approach to estimating the entropy of transformers' latent space, that allows for non-parametric analysis of representational structure. Our findings help shed light on what the learning process looks like in deep-learning models, and makes a case that intuitions from linguistics and cognitive science about what makes for a `good' representation may meaningfully transfer here.

\section{Acknowledgements}
We’d like to thank Ivan Titov for advice and discussion.
H Conklin was supported in part by the UKRI Centre for Doctoral Training in Natural Language
Processing, funded by the UKRI (grant EP/S022481/1) and the University of Edinburgh, School of
Informatics and School of Philosophy, Psychology \& Language Sciences.

\bibliographystyle{apacite}

\setlength{\bibleftmargin}{.125in}
\setlength{\bibindent}{-\bibleftmargin}

\bibliography{references, additional}

\end{document}